\documentclass{article}

    \PassOptionsToPackage{numbers, compress}{natbib}

\usepackage[final]{unireps_2024}

\usepackage[utf8]{inputenc} 
\usepackage[T1]{fontenc}    
\usepackage{hyperref}       
\usepackage{url}            
\usepackage{booktabs}       
\usepackage{amsfonts}       
\usepackage{nicefrac}       
\usepackage{microtype}      
\usepackage{xcolor}         

\usepackage{import}
\usepackage{subcaption}
\usepackage{amsmath}
\usepackage[]{graphicx}
\usepackage[ruled,vlined,linesnumbered]{algorithm2e}
\usepackage{multirow}       

\usepackage{cleveref} 
\usepackage{amsthm}

\newtheorem{prop}{Proposition}

\title{Improving OOD Generalization of Pre-trained Encoders via Aligned Embedding-Space Ensembles}

\author{
    Shuman Peng,\hspace{0.25ex} Arash Khoeini,\hspace{0.25ex} Sharan Vaswani,\hspace{0.25ex} Martin Ester \\
    School of Computing Science\\
    Simon Fraser University \\
    {\tt\small \{shumanp, akhoeini\}@sfu.ca } 
    {\tt\small vaswani.sharan@gmail.com }
    {\tt\small ester@cs.sfu.ca }
}

\begin{document}

\maketitle

\begin{abstract}
    The quality of self-supervised pre-trained embeddings on out-of-distribution (OOD) data is poor without fine-tuning. A straightforward and simple approach to improving the generalization of pre-trained representation to OOD data is the use of deep ensembles. However, obtaining an effective ensemble in the embedding space with only unlabeled data remains an unsolved problem. We first perform a theoretical analysis that reveals the relationship between individual hyperspherical embedding spaces in an ensemble. We then design a principled method to align these embedding spaces in an unsupervised manner. Experimental results on the MNIST dataset show that our embedding-space ensemble method improves pre-trained embedding quality on in-distribution and OOD data compared to single encoders. 
\end{abstract}

\section{Introduction}
Self-supervised learning techniques enable the pre-training of deep neural network (DNN) encoders on widely available unlabeled data.
These pre-trained encoders, once fine-tuned, are highly transferable to various downstream tasks \cite{he2020momentum,chen2020simple}.
However, a key challenge remains: without fine-tuning, the quality of pre-trained features is noticeably lower on out-of-distribution (OOD) data, impairing the performance of pre-trained models on subsequent OOD downstream tasks
\citep{kirchhof2023url}. This issue is particularly critical when the downstream task has insufficient data for fine-tuning, making it essential for the pre-trained encoders to generalize well to OOD data in a zero-shot manner without fine-tuning.

A straightforward approach to improve the generalizability of pre-trained representation quality to OOD data is the use of deep ensembles. Deep ensembles (DEs) \citep{lakshminarayanan2017DeepEnsemble}, consist of DNNs independently trained with different initializations and data orders (e.g. seeds). DEs have been shown to improve the predictive performance over single DNNs on both in-distribution (ID) and OOD data \citep{ovadia2019can, Arpit2021EnsembleOA}. 

Existing DE approaches typically aggregate models either in the predictive output space \citep{Arpit2021EnsembleOA, lakshminarayanan2017DeepEnsemble, ovadia2019can} (e.g., logits for classification) or in the weight space \citep{Wortsman2021RobustFO, rame2022diverse, rame2023model}. Aggregating in the predictive space confines the ensemble to a single task, e.g., classification with a fixed set of classes, and is inapplicable for self-supervised pre-trained encoders. In contrast, aggregating models in the weight space offers the flexibility to accommodate various tasks by discarding the predictive layer and retaining only the ensembled encoder. However, this approach sacrifices the interpretability that predictive-space ensembles provide, particularly in how individual model outputs are combined and unified. In essence, existing DE techniques exhibit an undesirable trade-off between interpretability and flexibility.

In this paper, we take a novel perspective to ensemble self-supervised pre-trained encoders for improved zero-shot OOD generalization. Our approach offers both the flexibility of weight-space ensembling and the interpretability of predictive-space ensembling. We propose an embedding-space ensemble, called \texttt{Ensemble-InfoNCE}, in which we aggregate the ensemble mean in the hyperspherical latent embedding space of encoders pre-trained using the widely used InfoNCE contrastive loss \citep{oord2018infonce, chen2020simple}.

\begin{figure}[t]
    \vspace{-3em}
    \centering
    \includegraphics[width=0.63\linewidth]{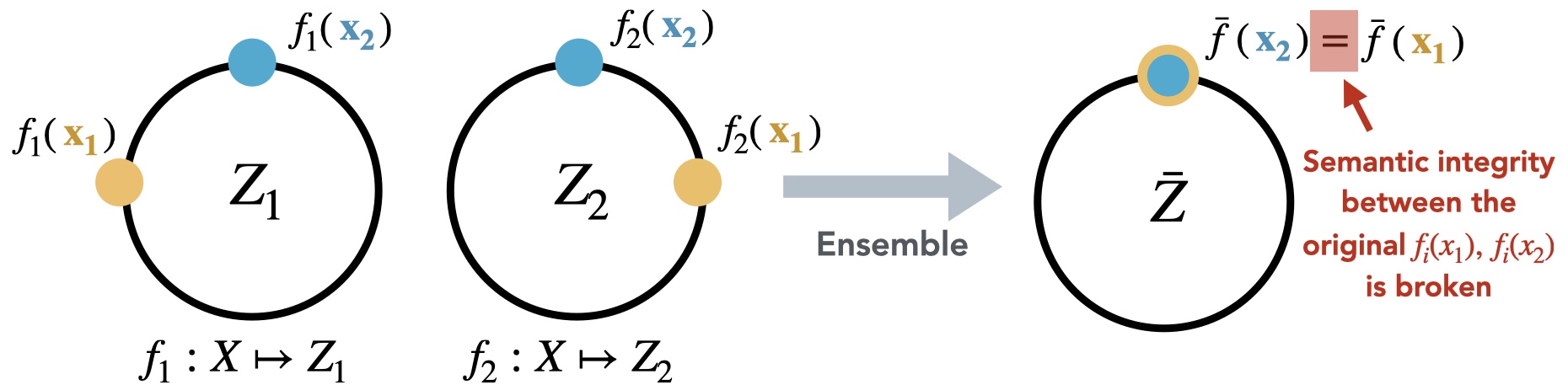}
    \vspace{-0.5em}
    \caption{Need for embedding alignment: The ensemble mean of two different embeddings (yellow, blue) in misaligned embedding spaces $Z_1, Z_2$ collapses to the same vector in $\bar{Z}$, although they have different semantic meanings. 
    }
    \label{fig:misaligned-ensemble-mean-toy-example}
    \vspace{-1em}
\end{figure}

Obtaining the ensemble mean of embedding vectors is less straightforward compared to taking the mean of predictive outputs or model weights. To take the ensemble mean of embeddings, the embedding spaces must be aligned such that the embeddings (produced by the different encoders) corresponding to semantically similar samples have a similar direction in the hyperspherical space. Taking the mean of misaligned embeddings that point in different directions can harm the semantic integrity of the embedding space (Figure \ref{fig:misaligned-ensemble-mean-toy-example}). 
Existing approaches align embeddings using class labels \citep{xu2023probabilistic}. However, aligning embeddings without access to labels remains an unsolved problem. To this end, we extend the theoretical results in \citep{zimmermann2021contrastive}, and use this to propose a principled unsupervised approach (referred to as \texttt{Ensemble-InfoNCE}) to align the embeddings.
Our theoretical results demonstrate that an ensemble of encoders with aligned embedding spaces recovers the correct (ground truth) embeddings.
Finally, we experimentally show improved pre-trained embedding quality on in-distribution (ID) and OOD data for \texttt{Ensemble-InfoNCE} compared to single InfoNCE encoders on the MNIST dataset.

\section{Background and preliminaries}

\paragraph{InfoNCE loss.}
\label{sec:background-infonce}
\vspace{-0.5em}
We consider encoders pre-trained using the InfoNCE loss, a widely used contrastive (self-supervised) learning objective \citep{oord2018infonce, bachman2019learning, chen2020simple, he2020momentum, tian2020contrastive, wang2020understanding}. These encoders $f$ map the input space $\mathcal{X}$ to an $L_2$-normalized unit-hyperspherical embedding space $\mathcal{Z} = \mathbb{S}^{D-1}$. Encoders trained with the InfoNCE loss (\ref{eq:contrastive_loss}) have the desirable property of mapping semantically similar pairs of samples close together in the embedding space, while also ensuring that dissimilar samples are mapped far apart.

\vspace{-1em}
\paragraph{InfoNCE-trained encoders recover the correct latents.}
\label{sec:background-correct-latents}

\citet{zimmermann2021contrastive} theoretically demonstrate that an encoder $f$ minizing the contrastive loss in (\ref{eq:contrastive_loss}) recovers the ground truth latents $z \in \mathcal{Z}$ up to orthogonal transformations. Specifically, $z_1^\top z_2 = h(z_1)^\top h(z_2)$, where $z_1, z_2$ are ground truth latents, and $h = f \circ g$ composes of encoder $f$ and a generative process $g: \mathcal{Z} \mapsto \mathcal{X}$.\footnote{We refer interested readers to Theorem 2 of \citep{zimmermann2021contrastive}.}

\section{Embedding-space ensembles via unsupervised alignment}

In this section, we introduce, to the best of our knowledge, the first approach of ensembling self-supervised pre-trained encoders in the embedding space, which offers the interpretability of predictive-space ensembles and the flexibility of weight-space ensembles.
Our embedding-space ensemble approach, referred to as \texttt{Ensemble-InfoNCE}, produces mean embedding vectors $\bar{z} = \bar{f}(x) \in \mathcal{Z}$ for a given ensemble of $M>1$ encoders $\{f_i: \mathcal{X} \mapsto \mathcal{Z}_i\}_{i=1}^M$ pre-trained using the InfoNCE loss (\ref{eq:contrastive_loss}). Each $\mathcal{Z}_i$ is a unit-hyperspherical embedding space $\mathcal{Z}_i = \mathbb{S}^{D-1}$.
Before taking the ensemble mean, we must first \textit{align} the embedding spaces $\{\mathcal{Z}_i\}_{i=1}^M$ so that each of the $M$ embeddings $\{f_i(x)\}_{i=1}^M$ for the same input $x$ points in a similar direction on a hypersphere. However, aligning embedding spaces without labels in an unsupervised manner remains a challenging problem.

To tackle the challenge of performing unsupervised embedding space alignment, we first conduct a theoretical analysis and reveal a critical orthogonality relationship between different embedding spaces (Section \ref{sec:theoretical-analysis-z1-Rz2}). 
This relationship allows us to extend the guarantees on the recovery of the correct latents\footnote{In this paper, we use the terms "latents", "features", and "embeddings" interchangeably.} from single encoders to an ensemble. Furthermore, this relationship enables us to align the embedding spaces by recovering the orthogonal transformation matrix and forms the basis of our unsupervised embedding space alignment approach (Section \ref{sec:method-alignment}). Finally, the aligned ensemble embeddings are aggregated using the Karcher Mean algorithm \citep{straub2015dirichlet} (Section \ref{sec:method-ensemble-mean}).


\subsection{Theoretical analysis}
\label{sec:theoretical-analysis-z1-Rz2}

\paragraph{An encoder's latent space is an orthogonal transformation of another encoder's.}
Extending the main theoretical result of \cite{zimmermann2021contrastive} (Section \ref{sec:background-correct-latents}) from single encoders to an ensemble of these encoders, we reveal that an encoder's embedding space is an orthogonal transformation of another encoder's. This is formally stated below in Proposition \ref{corollary:f1-f2-rotation} with the corresponding proof in \Cref{sec:supp-proof-for-theoretical-analysis-sec}.

\vspace{-1em}
\paragraph{Assumption.}
For an ensemble of encoders $f_1, f_2$ trained on the same data $D = \{x_i\}_{i=1}^N$ with different random seeds, we \textit{assume} that $f_1(x) = R_1 z$ and $f_2(x) = R_2 z$. In other words, both $f_1$ and $f_2$ recover the correct (ground truth) latents $z$ up to different orthogonal transformations $R_1, R_2$.\footnote{For our theoretical analysis, we consider the case of an ensemble of $M=2$ encoders for simplicity. The results can be extended to $M>2$ encoders.}

\begin{prop}[Orthogonal transformation relationship]
\label{corollary:f1-f2-rotation}
    Under the above assumption, $f_1$ and $f_2$ learn the same latents up to an orthogonal transformation $R$, that is, $f_1(x) = R f_2(x)$.
\end{prop}

\vspace{-0.8em}
\paragraph{An ensemble of aligned embeddings recovers the correct latents.}

Based on Proposition \ref{corollary:f1-f2-rotation}, in Proposition \ref{prop:ensemble-recovers-correct-latents} we generalize the theoretical guarantee on the correctness of the learned latents for each ensemble member to the ensemble as a whole. 
The proof is provided in \Cref{sec:supp-proof-for-theoretical-analysis-sec}. 
\begin{prop}[Ensemble recovers correct latents]
\label{prop:ensemble-recovers-correct-latents}
    The ensemble mean $\bar{f}(x)$ of aligned embeddings $f_1(x)$ and $R f_2(x)$ are the correct latents $z$ up to orthogonal transformation $R_1$, that is, $\bar{f}(x) = R_1 z$. 
\end{prop}

\vspace{-0.8em}
\paragraph{Approximate orthogonal relationship on real-world data.}
\label{sec:approx-orthogonal-transformation}

With real-world data, there may not exist an orthogonal transformation that perfectly aligns \textit{all} corresponding embedding vectors of the same input between different embedding spaces. This discrepancy arises due to violations of the data generation and modeling assumptions.
Under such violated conditions, \citet{zimmermann2021contrastive} demonstrated that these encoders still recover the true latents to a moderate to high degree. 
By extending this result to our problem setting, we infer that the relationship between encoders $f_i$ and $f_j$ can be approximated as $f_i(x) \approx R f_j(x)$ for all $x \in \mathcal{X}$. 
In deep ensembles, this approximate orthogonal transformation relationship preserves some degree of diversity between embedding spaces, preventing the ensemble from collapsing into a single model.

\subsection{Unsupervised embedding space alignment via learning orthogonal matrices}
\label{sec:method-alignment}
 
Our goal is to align the $M$ hyperspherical embedding spaces $Z_1, ..., Z_M$ induced by encoders $\{f_i\}_{i=1}^M$ so that the same in-distribution pre-training sample is mapped to embeddings that have a similar direction across $\{Z_i\}_{i=1}^M$. From \Cref{sec:approx-orthogonal-transformation}, we know that $f_i(x) \approx R f_j(x)$ for all $x \in \mathcal{X}$, which means that learning the orthogonal transformation matrix $R$ would naturally align $f_j(x)$ with $f_i(x)$. 
We note that $R$ does not need to be strictly orthogonal for aligning the embedding spaces. To learn $R$ in a $D$-dimensional embedding space, we use a single-layer neural network with $D$ input and output nodes. The $D \times D$ weight matrix within this single layer neural network represents the orthogonal transformation matrix $R$ that we want to learn. 

To align the embedding spaces, we randomly select one \textbf{anchor} encoder from the set of $M$ encoders to align the remaining $M-1$ encoders. To align each $Z_j$ with the anchor embedding space $Z_i$, we propose an objective function that enforces $R$ to be as close to orthogonal as possible by imposing orthogonality as a soft constraint, while simultaneously maximizing the alignment between pairs of embeddings: 

\vspace{-6ex}
        \begin{align}
            \mathcal{L}_\text{align} & = \underset{R \in \mathbb{R}^{D \times D}}{\arg\min} \; \frac{1}{N} \underset{n=1}{\overset{N}{\sum}} \; d (f_i(x_n), R f_j(x_n)) + \lambda \|R^T R - I_D\|_F^2
        \end{align}
where $N$ denotes the number of samples, $\lambda$ is a hyperparameter that controls the strength of the orthogonality constraint, and $d(.,.)$ is a function that quantifies the discrepancy between pairs of vectors. 
Given that the embedding vectors reside on the surface of a unit sphere, we use the geodesic distance, which measures the shortest path between two points on a Riemannian manifold and accounts for the spherical geometry. 
The objective function then becomes:
\vspace{-1ex}
        \begin{align}
            \mathcal{L}_\text{align} & = \underset{R \in \mathbb{R}^{D \times D}}{\arg\min} \; \frac{1}{N} \underset{n=1}{\overset{N}{\sum}} \; \arccos (f_i(x_n), R f_j(x_n)) + \lambda \|R^T R - I_D\|_F^2
        \end{align}

\subsection{Embedding mean in an ensemble}
\label{sec:method-ensemble-mean}

An ensemble of $M$ pre-trained encoders with aligned embedding spaces produces a mean embedding vector for each sample, i.e., $\bar{z} = \bar{f}(x) = \text{mean}(f_1(x), R_2f_2(x),..., R_Mf_M(x))$. 
Given the hyperspherical nature of the embedding spaces, we apply the Karcher Mean algorithm from \citep{straub2015dirichlet} to compute meaningful mean vectors on the surface of a sphere. The algorithm projects hyperspherical data points onto a linear tangent space, calculates the mean in this tangent space, and then projects the result back onto the sphere. This process iterates until the mean in the tangent space approaches a near-zero norm, indicating that the mean has converged.\footnote{Details of the Karcher Mean algorithm can be found in \citep{straub2015dirichlet}.}

\section{Experiments}
\vspace{-1ex}

\begin{figure}[t]
    \vspace{-4em}
    \centering
    \begin{subfigure}[b]{0.45\textwidth}
        \centering
        \includegraphics[width=\textwidth, height=0.65\textwidth]{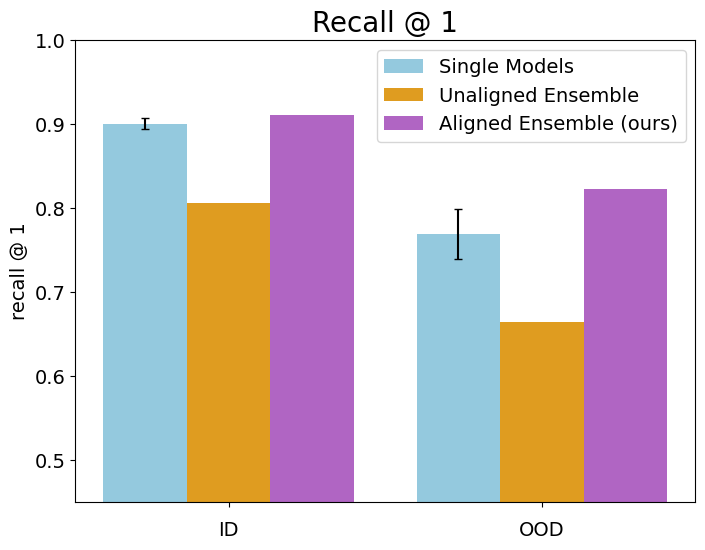} 
    \end{subfigure}
    \hfill
    \begin{subfigure}[b]{0.45\textwidth}
        \centering
        \includegraphics[width=\textwidth, height=0.65\textwidth]{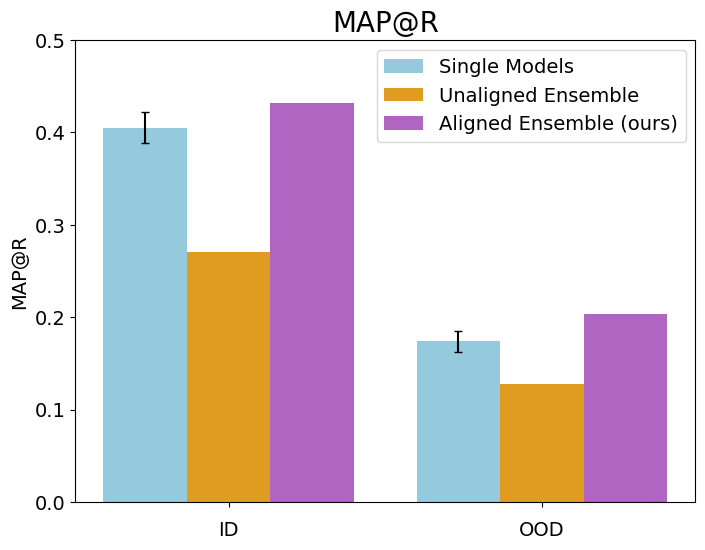} 
    \end{subfigure}
    \vspace{-1.5ex}
    \caption{Comparing embedding qualities of single models (blue), an ensemble of unaligned embedding spaces (orange), and an ensemble of aligned embedding spaces (purple) in the ID and OOD settings. Recall@1 and MAP@R are presented. Higher values indicate better performance. The mean and standard deviation (error bars) of the performance metrics are reported for the 5 single models. The ensembles do not have standard deviation since all 5 models are combined into one.}
    \label{fig:main-results-r1-mapr-color}
    \vspace{-1em}
\end{figure}

\paragraph{Dataset and training setup} 
We use the MNIST dataset \citep{lecun1998mnist}. For ID evaluation, the test set is used as is. For OOD evaluation, each test sample is randomly colored. Since random coloring was not applied during pre-training, colored versions of the images are considered OOD. A total of $M=5$ encoders are trained for the ensemble, as this ensemble size has been shown to be sufficient to produce good results \citep{ovadia2019can}. Details of the dataset, model architecture and training are provided in \Cref{sec:supp-training-details} and \ref{sec:supp-dataset-details}.
\vspace{-1em}

\paragraph{Evaluation metrics} 
In line with the representation learning literature, we assess the quality of embeddings using the \textbf{Recall at 1} (R@1) and \textbf{Mean Average Precision at R} (MAP@R) metrics \citep{kirchhof2023url, musgrave2020metric}. 
R@1 measures the semantic quality of the embedding space by verifying if each embedding's nearest neighbor belongs to the same class. 
MAP@R evaluates the proportion of each embedding's $R$ nearest neighbors that belong to the same class, while accounting for the ranking of correct retrievals \citep{musgrave2020metric}. $R$ is set to the total number of samples in a class \citep{musgrave2020metric}.

\subsection{Experimental results}
\vspace{-1ex}
Figure \ref{fig:main-results-r1-mapr-color} shows that our \texttt{Ensemble-InfoNCE} model with aligned embedding spaces (shown in purple) improves the quality of embeddings over single models and unaligned ensembles in both the in-distribution (ID) and out-of-distribution (OOD) settings. The embedding quality improvement achieved by our method is more pronounced in the OOD setting, with a $6.99\%$ improvement over the mean R@1 of the $M=5$ single models and a $17.38\%$ improvement over the mean MAP@R of single models. 
Our results also highlight the importance of aligned embedding spaces for ensembles. Taking an ensemble of misaligned embedding spaces consistently hurts the embedding quality, achieving lower values of R@1 and MAP@R than a single model for both ID and OOD settings.
Table \ref{tab:r1-mapr-single-models-vs-embedding-space-ensembles} provides the numerical values for Figure \ref{fig:main-results-r1-mapr-color}, and additional results are provided in \Cref{sec:supp-additional-results}.


\section{Conclusion}
\vspace{-2ex}
We improved the generalizability of pre-trained encoders to OOD data by taking an ensemble of encoders in the embedding space. We constructed embedding-space ensembles by effectively aligning the individual embedding spaces. Preliminary experiments on the MNIST dataset demonstrate that our aligned embedding-space ensemble significantly enhances the OOD embedding quality compared to individual models. 
In the future, we will focus on scaling our method to larger datasets, such as ImageNet, and incorporating OOD data from real datasets \citep{deng2009imagenet, krause2013cars, wah2011cub, oh2016stanfordonlineproducts}.

\begin{ack}
This research was supported by the NSERC Discovery Grant. We would like to thank Dr. Ke Li for his early feedback. We are grateful to Shichong Peng for providing insightful discussions and thorough feedback throughout the course of this research. We also appreciate the anonymous reviewers for their constructive feedback and suggestions. 
\end{ack}



\bibliographystyle{plainnat}
\bibliography{main}

\newpage
\appendix

\section{Appendix / supplemental material}

\subsection{The InfoNCE contrastive loss}

The InfoNCE loss is defined as
\begin{align} \label{eq:contrastive_loss}
        &L_{\text{contr}}(f; \tau, M) \quad := 
        \underset{\substack{
            (x, \;x^+) \sim p_\mathsf{pos} \\
            \{x^-_i\}_{i=1}^M \overset{\text{i.i.d.}}{\sim} p_\mathsf{data}
        }}{\mathbb{E}} \left[\, {- \log \frac{e^{f(x)^{\mathsf{T}} f(x^+) / \tau }}{e^{f(x)^{\mathsf{T}} f(x^+) / \tau } + \sum\limits_{i=1}^M e^{f(x)^{\mathsf{T}} f(x^-_i) / \tau }}}\,\right], 
    \end{align}
where pairs $(x, x^+)$ are drawn of the distribution of positive samples $p_{\text{pos}}$, and $M>0$ negative samples are drawn from the distribution of all observations $p_{\text{data}}$. 

\subsection{Proofs for Section \ref{sec:theoretical-analysis-z1-Rz2}}
\label{sec:supp-proof-for-theoretical-analysis-sec}

\paragraph{Proposition \ref{corollary:f1-f2-rotation}}
\textit{Under the above assumption, $f_1$ and $f_2$ learn the same latents up to an orthogonal transformation $R$, that is, $f_1(x) = R f_2(x)$.}

\begin{proof}
Let $f_1(x) = R_1 z$ and $f_2(x) = R_2 z$ where $R_1, R_2$ are orthogonal matrices, i.e., $R_1^T = R_1^{-1}, R_2^T = R_2^{-1}$, we have
\begin{align*}
    &\Rightarrow R_1^{-1} f_1(x) = z \quad \text{and} \quad R_2^{-1}f_2(x) = z \\
    &\Rightarrow R_1^{-1} f_1(x) = R_2^{-1} f_2(x) \\
    &\Rightarrow R_1 R_1^{-1} f_1(x) = R_1 R_2^{-1} f_2(x) \\
    &\Rightarrow f_1(x) = Rf_2(x) \quad \text{(Letting} \; R = R_1 R_2^{-1} \text{)}
\end{align*}
Since  $R_1$ and $R_2^{-1}$ are both orthogonal matrices, their product $R$ is also an orthogonal matrix, i.e., $R^T = R^{-1}$. Therefore, $f_1$ and $f_2$ learns the same latents up to an orthogonal transformation $R$. 
\end{proof}

\paragraph{Proposition \ref{prop:ensemble-recovers-correct-latents}}\hspace{-0.5em}(The ensemble also recovers the correct latents)\textbf{.}
\textit{The ensemble mean $\bar{f}(x)$ of aligned embeddings $f_1(x)$ and $R f_2(x)$ are the correct latents $z$ up to orthogonal rotation $R_1$, that is, $\bar{f}(x) = R_1 z$. }

\begin{proof}
Let us denote the ensemble mean as $\bar{f}(x) = \text{mean}(f_1(x), Rf_2(x))$, where $\text{mean}(.)$ is a general notion of the mean, which can be the arithmetic mean in Euclidean spaces or the Karcher Mean in Reimmanian manifolds. For the simplicity of this proof, we will use the arithmetic mean, but the results also apply to the Karcher Mean. 

Defining $\bar{f}(x)$ using the arithmetic mean, we have:

\begin{align}
    \bar{f}(x) & = \frac{1}{2} [ f_1(x) + Rf_2(x) ]
\end{align}

Since $f_1(x) = Rf_2(x)$, we have:

\begin{align}
    &\bar{f}(x) = \frac{1}{2} [ f_1(x) + f_1(x) ] = f_1(x) = R_1 z 
\end{align}

\end{proof}

\subsection{Additional implementation and training details}
\label{sec:supp-training-details}

\paragraph{Contrastive pre-training architecture} 
For contrastive pre-trained encoders, two convolution blocks with max-pooling and ReLU activations are used, with a linear layer attached at the end to project the embeddings down to $D=8$ dimensions. The first convolution block consists of (1) a Conv2d with 3 input channels, 16 output channels, a kernel size of 5, stride of 1, and padding of 2; (2) max-pooling with kernel size 2; (3) a ReLU activation; and finally (4) a dropout layer with dropout rate $p$ (we used $p=0.25$ in our experiments). Similarly, the second convolution block consists of identical components, except with a Conv2d that consists of 16 input channels and 32 output channels.

\paragraph{Supervised contrastive pre-training}
For part of our experiments (Figures \ref{fig:results-r1-mapr-color-supervised} and \ref{fig:results-r1-mapr-crop-supervised}), we follow \cite{kirchhof2023probabilistic} and \textbf{use class labels to generate positive and negative pairs}, which better preserves the theoretical assumptions and guarantees for encoders trained with the InfoNCE loss \citep{zimmermann2021contrastive}. Positive pairs consist of samples from the same class, while negative pairs are from different classes. We refer to this approach as \textit{supervised contrastive pre-training}.
Each encoder is contrastive trained for 2000 epochs using different random seeds ($10, 11, 12, 13, 14$) and weight initializations. A batch size of 128 is used, and each sample is paired with 16 negative samples, following \cite{kirchhof2023probabilistic}. The learning rate is set to $0.01$ and the AdamW optimizer \citep{loshchilov2017decoupled} is used.

\paragraph{Unsupervised contrastive pre-training}
For unsupervised contrastive pre-training, which is the conventional contrastive pre-training approach (also used for Figures \ref{fig:main-results-r1-mapr-color} and \ref{fig:results-r1-mapr-crop-unsupervised} and Table \ref{tab:r1-mapr-single-models-vs-embedding-space-ensembles}), we \textbf{apply random rotations ($\pm 30$ degrees) to generate positive pairs}. Positive pairs are created by taking two randomly rotated views of the same sample, while negative pairs are formed from randomly rotated views of different samples. We also experimented with using random cropping to generate positive pairs, but found that this resulted in unstable model training. This instability is likely due to the nature of the MNIST images, which consist of white digits on a black background. Images that are cropped to include too much background and insufficient detail of the digit can result in distinct samples being mistakenly identified as similar.
Each encoder is trained for 20 epochs using different random seeds ($10, 11, 12, 13, 14$) and weight initializations. We use a batch size of 1024 and a learning rate of 0.1 with the LAMB optimizer \citep{you2019large}, which is better suited for larger batch sizes.

\paragraph{Unsupervised embedding space alignment}
For unsupervised embedding space alignment, a linear layer with $D$ input and output dimensions is used. 
To align the \textbf{supervised contrastive pre-trained encoders}, the linear alignment layer layer is trained for 20 epochs with a learning rate of $0.1$, and an orthogonality regularization factor $\lambda = 0.5$ is applied. To align the \textbf{unsupervised contrastive pre-trained encoders}, which exhibit lower degrees of orthogonality compared to the supervised ones (as expected due to the violation of theoretical assumptions regarding the conditional distribution used to generate positive pairs \cite{zimmermann2021contrastive}), the linear layer is trained for 20 epochs with a learning rate of 0.1 and $\lambda \in \{0.1, 0.3, 0.5\}$ was applied. A lower
$\lambda$ relaxes the orthogonality constraint for encoders that have slightly weaker orthogonal relationships. In both cases, the linear layer weights were optimized using stochastic gradient descent (SGD).

\paragraph{Computing resources}
We used a single RTX 3090 GPU for our experiments. For MNIST scale experiments, any GPU with more than 8GB of VRAM would be sufficient.

\subsection{Additional dataset details}
\label{sec:supp-dataset-details}

We converted single-channel grayscale MNIST images to three-channel black-and-white images. The training set is used to perform contrastive pre-training of the encoders and to align the embedding spaces. 
The test set is used for ID and OOD evaluation of the pre-trained encoders. 

\subsubsection{Data for OOD evaluation}
\label{sec:ood-evaluation-data}
\begin{figure}[t]
    \centering
    \begin{subfigure}[b]{0.45\textwidth}
        \centering
        \includegraphics[width=\textwidth]{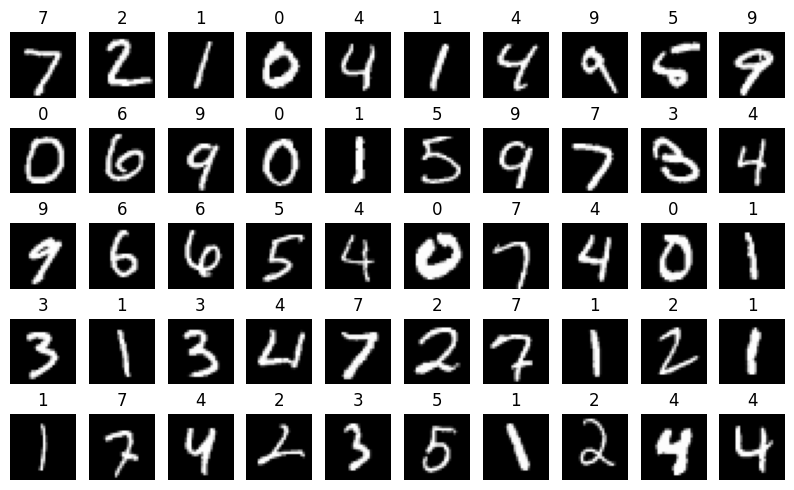} 
        \caption{Original (ID) MNIST}
        \label{fig:original-mnist}
    \end{subfigure}
    \hfill
    \begin{subfigure}[b]{0.45\textwidth}
        \centering
        \includegraphics[width=\textwidth]{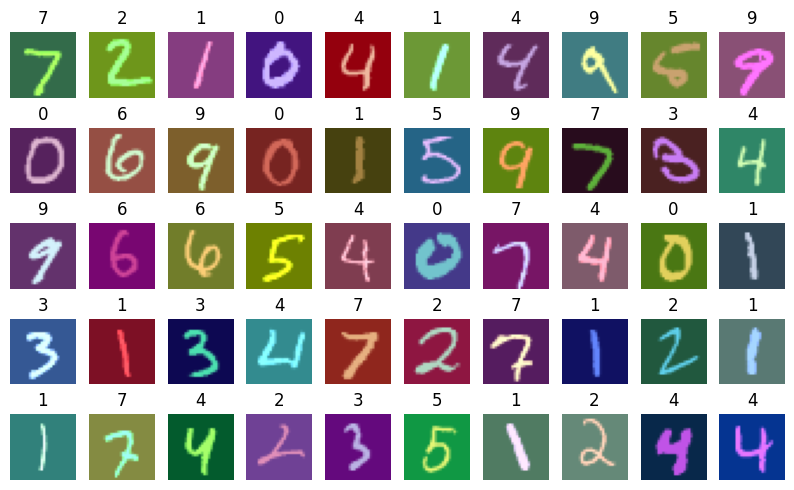} 
        \caption{Colored (OOD) MNIST}
        \label{fig:colored-mnist}
    \end{subfigure}
    \caption{For in-distribution (ID) evaluation, images like those in (a) were used. For out-of-distribution (OOD) evaluation, images like those in (b) were used.}
    \label{fig:colored-mnist-ood}
\end{figure}

\paragraph{Colored version}
We randomly colored the images in the MNIST test set to create an OOD evaluation set. Since only black-and-white images were used during pre-training, colored versions of the images are considered OOD compared to the original images. Colored versions of the images are illustrated in \Cref{fig:colored-mnist}.

\paragraph{Cropped version}
For further OOD evaluation, each test sample is randomly cropped to \texttt{crop\_size} $\sim \text{Unif}([0.25, 1])$ percent of their original size, following \citep{kirchhof2023probabilistic, kirchhof2023url}. Since no random cropping was applied during pre-training, cropped versions of the images are considered OOD compared to the original images.

\subsection{Additional results}
\label{sec:supp-additional-results}

\subsubsection{Supervised contrastive pre-training with colored OOD images}
\begin{figure}[t]
    \centering
    \begin{subfigure}[b]{0.45\textwidth}
        \centering
        \includegraphics[width=\textwidth, height=0.65\textwidth]{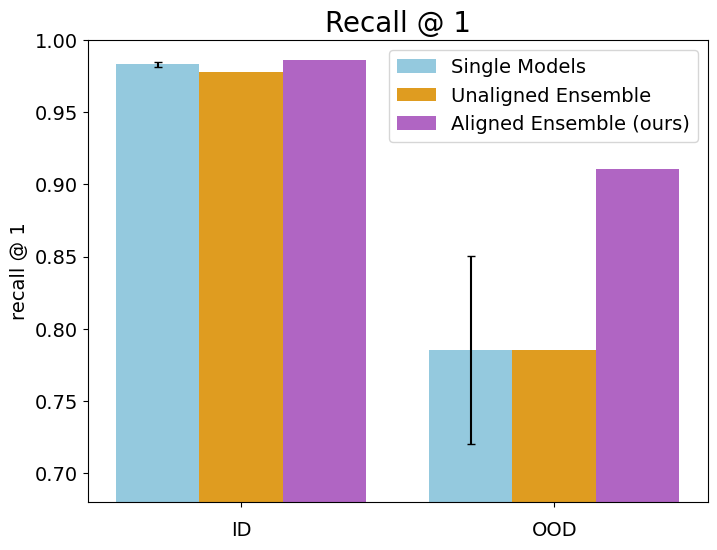} 
    \end{subfigure}
    \hfill
    \begin{subfigure}[b]{0.45\textwidth}
        \centering
        \includegraphics[width=\textwidth, height=0.65\textwidth]{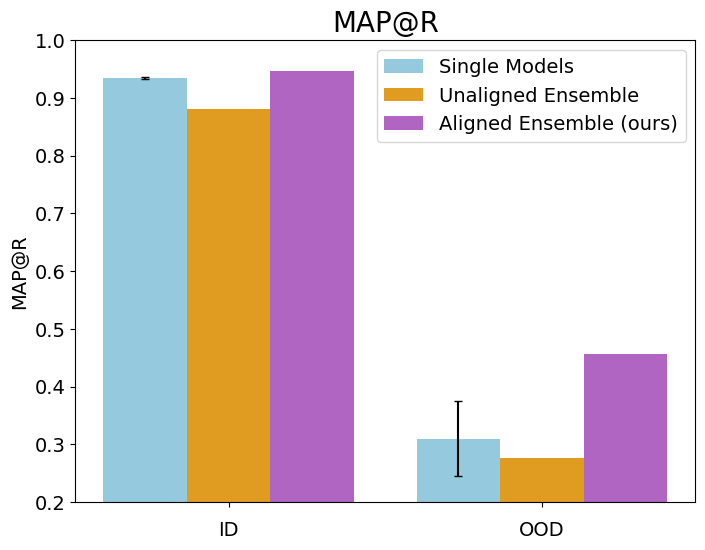} 
    \end{subfigure}
    \caption{Supervised contrastive pre-training with Colored MNIST as OOD evaluation data. Comparing embedding qualities of single models (blue), an ensemble of unaligned embedding spaces (orange), and an ensemble of aligned embedding spaces (purple) in the ID and OOD settings. Recall@1 and MAP@R are presented. Higher values indicate better performance. The mean and standard deviation (error bars) of the performance metrics are reported for the 5 single models. The ensembles do not have standard deviation since all 5 models are combined into one.}
    \label{fig:results-r1-mapr-color-supervised}
\end{figure}

Figure \ref{fig:results-r1-mapr-color-supervised} compares the embedding qualities of single models and ensembles of aligned and unaligned embedding spaces. The $M=5$ models are trained using the supervised contrastive pre-training procedure discussed in \Cref{sec:supp-training-details}. OOD evaluation is performed on Colored MNIST images (\Cref{sec:ood-evaluation-data}). 

\subsubsection{Supervised contrastive pre-training with cropped OOD images}
\begin{figure}[!h]
    \centering
    \begin{subfigure}[b]{0.45\textwidth}
        \centering
        \includegraphics[width=\textwidth, height=0.65\textwidth]{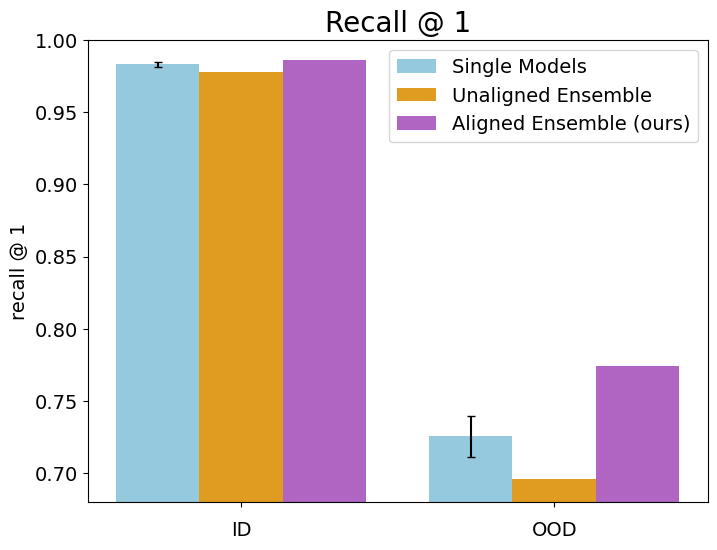} 
        \label{fig:r1-id-ood}
    \end{subfigure}
    \hfill
    \begin{subfigure}[b]{0.45\textwidth}
        \centering
        \includegraphics[width=\textwidth, height=0.65\textwidth]{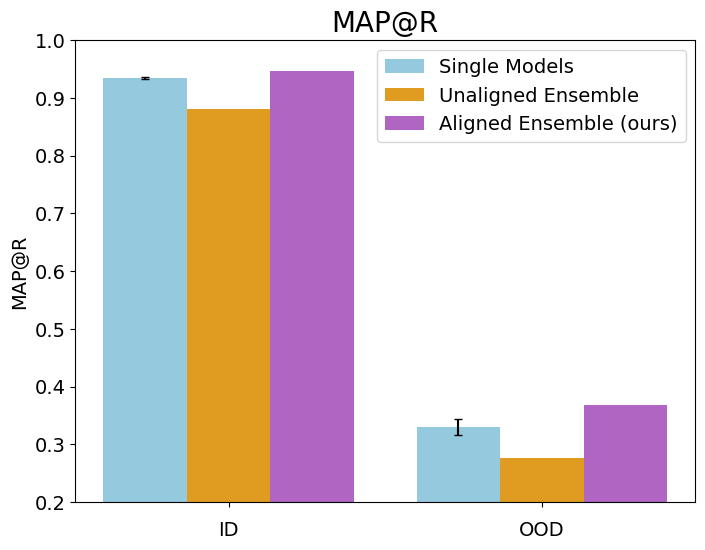} 
        \label{fig:mapr-id-ood}
    \end{subfigure}
    \vspace{-1.5em}
    \caption{Supervised contrastive pre-training with Cropped MNIST as OOD evaluation data. Comparing embedding qualities of single models (blue), an ensemble of unaligned embedding spaces (orange), and an ensemble of aligned embedding spaces (purple) in the ID and OOD settings. Recall@1 and MAP@R are presented. Higher values indicate better performance. The mean and standard deviation (error bars) of the performance metrics are reported for the 5 single models. The ensembles do not have standard deviation since all 5 models are combined into one.}
    \label{fig:results-r1-mapr-crop-supervised}
\end{figure}

Figure \ref{fig:results-r1-mapr-crop-supervised} compares the embedding qualities of single models and ensembles of aligned and unaligned embedding spaces. The $M=5$ models are trained using the supervised contrastive pre-training procedure discussed in \Cref{sec:supp-training-details}. Cropped MNIST (\ref{sec:ood-evaluation-data}) images are used as OOD evaluation data. 

\subsubsection{Unsupervised contrastive pre-training with cropped OOD images}
\begin{figure}[t]
    \centering
    \begin{subfigure}[b]{0.45\textwidth}
        \centering
        \includegraphics[width=\textwidth, height=0.65\textwidth]{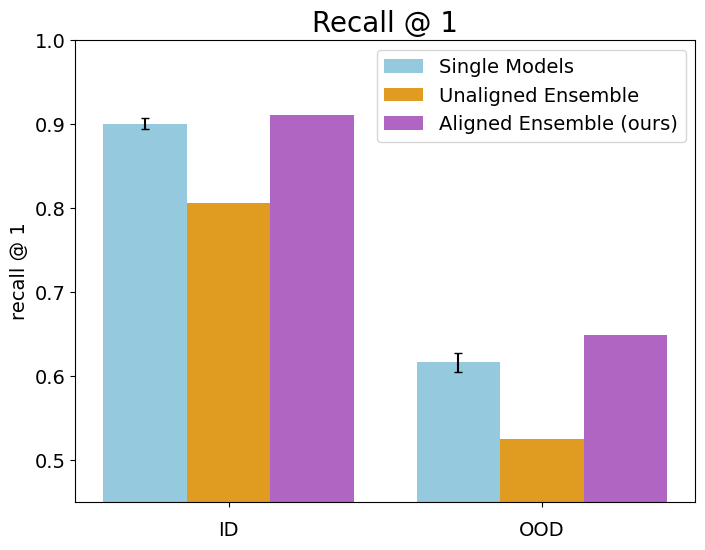} 
    \end{subfigure}
    \hfill
    \begin{subfigure}[b]{0.45\textwidth}
        \centering
        \includegraphics[width=\textwidth, height=0.65\textwidth]{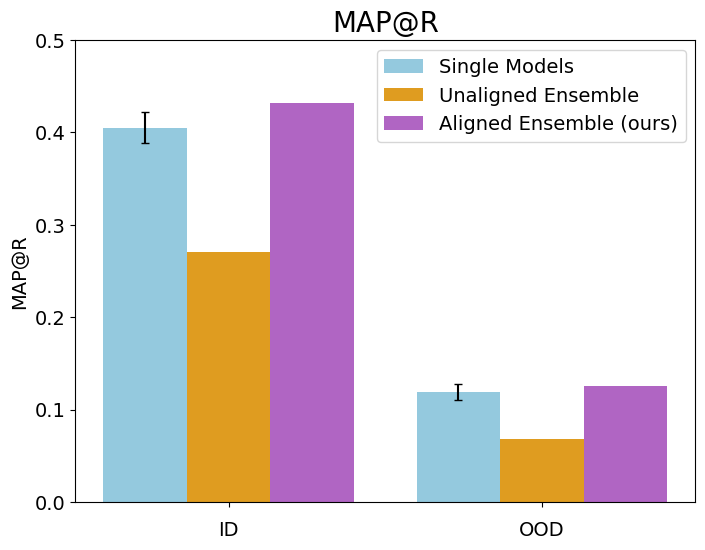} 
    \end{subfigure}
    \caption{Unsupervised contrastive pre-training with Cropped MNIST as OOD evaluation data. Comparing embedding qualities of single models (blue), an ensemble of unaligned embedding spaces (orange), and an ensemble of aligned embedding spaces (purple) in the ID and OOD settings. Recall@1 and MAP@R are presented. Higher values indicate better performance. The mean and standard deviation (error bars) of the performance metrics are reported for the 5 single models. The ensembles do not have standard deviation since all 5 models are combined into one.}
    \label{fig:results-r1-mapr-crop-unsupervised}
\end{figure}

Figure \ref{fig:results-r1-mapr-crop-unsupervised} compares the embedding qualities of single models and ensembles of aligned and unaligned embedding spaces. The $M=5$ models are trained using the unsupervised contrastive pre-training procedure discussed in \ref{sec:supp-training-details}. Cropped MNIST (\ref{sec:ood-evaluation-data}) images are used as OOD evaluation data. 

\subsubsection{Comparing embedding-space ensemble with weight-space ensemble}

\paragraph{Weight-space ensemble baselines}
We create two types of weight-space ensembles: \textbf{weight-space ensemble (WSE)} and \textbf{weight-space ensemble* (WSE*)}. For WSE, we combine the $M$ independently contrastive pre-trained encoders -- each with different initializations but otherwise identical hyperparameters -- by taking the uniform mean of their weights. WSE is directly comparable to our embedding-space ensembles, as it uses the same set of single models. However, effective weight averaging requires that the models be trained from the same initialization, but with different hyperparameters sampled from a \textit{mild search space}, which were only provided for ResNet50, to ensure the weights are averageable \citep{rame2022diverse}. To satisfy these conditions for WSE*, we follow a similar approach by weight-averaging single models trained with carefully selected hyperparameters. To train the single MNIST models (non-ResNet50 architecture) we adjust the learning rate from $0.1$ by adding values in $\{0.00001, 0.00003, 0.00005\}$, and randomly sample the dropout rate from $\{0.25, 0.3\}$. Finally, WSE* is formed by taking the uniform mean of the resulting $M$ models.

\begin{figure}[t]
    \centering
    \begin{subfigure}[b]{0.45\textwidth}
        \centering
        \includegraphics[width=\textwidth, height=0.75\textwidth]{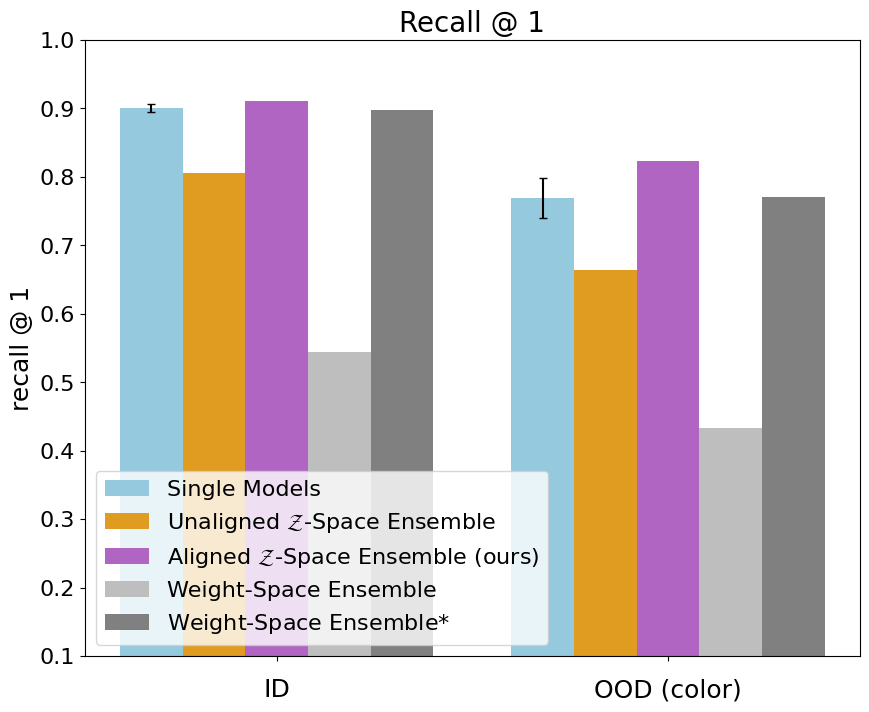} 
    \end{subfigure}
    \hfill
    \begin{subfigure}[b]{0.45\textwidth}
        \centering
        \includegraphics[width=\textwidth, height=0.75\textwidth]{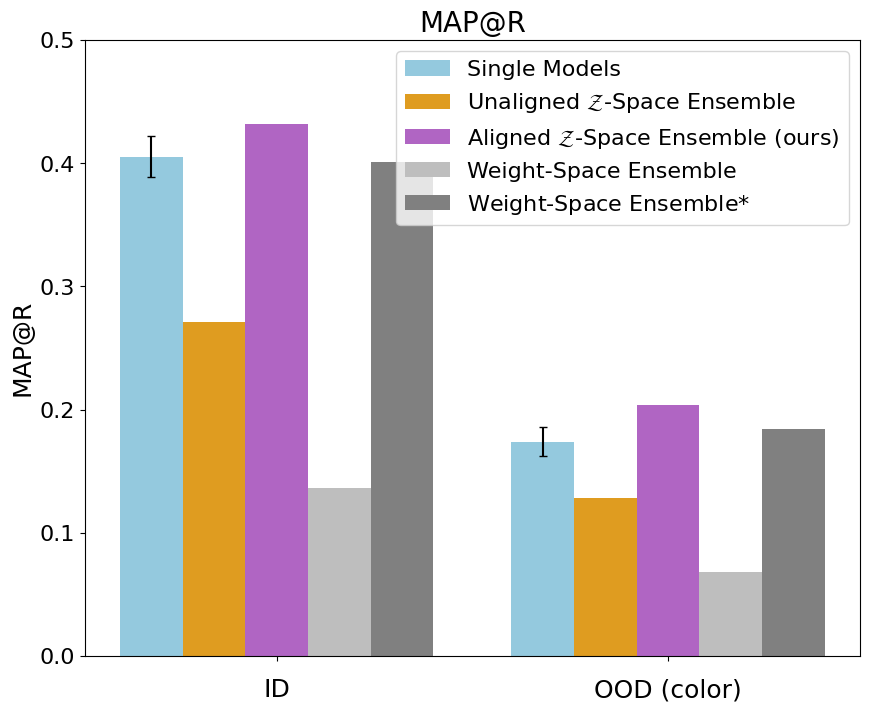} 
    \end{subfigure}
    \caption{Comparing embedding qualities of single models (blue), an ensemble of unaligned embedding spaces (orange), an ensemble of aligned embedding spaces (purple), a straightforward weight-space ensemble (WSE) (light grey), and a more constructed crafted weight-space ensemble (WSE*) \citep{rame2022diverse} (dark grey) in the ID and OOD settings. Recall@1 and MAP@R are presented. Higher values indicate better performance. The mean and standard deviation (error bars) of the performance metrics are reported for the 5 single models. The ensembles do not have standard deviation since all 5 models are combined into one. These plots show the same results as Figure \ref{fig:main-results-r1-mapr-color}, but with the addition of the weight-space ensemble baselines.}
    \label{fig:results-r1-mapr-color-unsupervised-diwa}
\end{figure}

\begin{figure}[t]
    \centering
    \begin{subfigure}[b]{0.45\textwidth}
        \centering
        \includegraphics[width=\textwidth, height=0.75\textwidth]{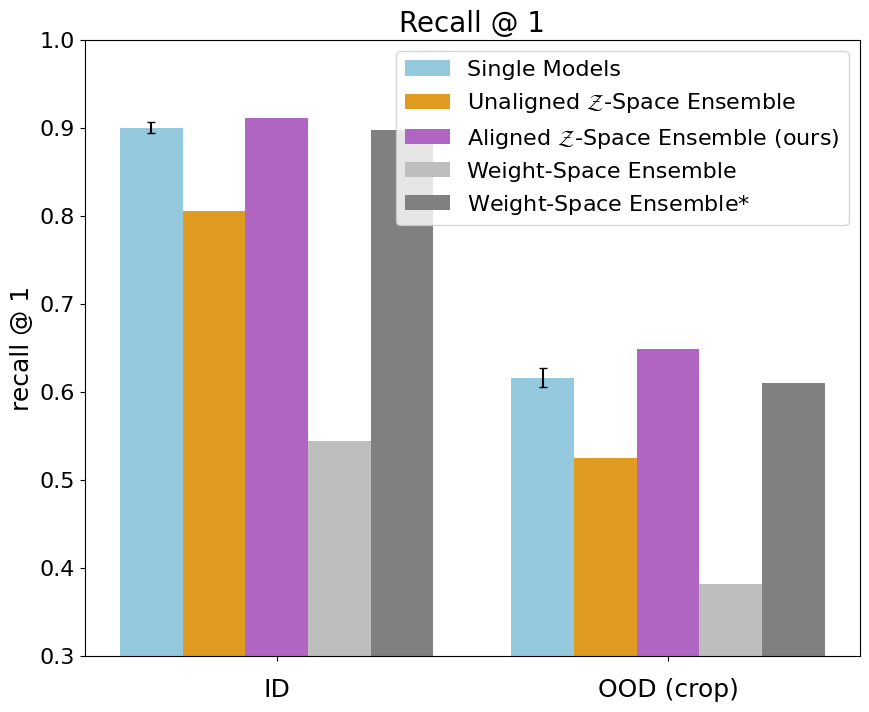} 
    \end{subfigure}
    \hfill
    \begin{subfigure}[b]{0.45\textwidth}
        \centering
        \includegraphics[width=\textwidth, height=0.75\textwidth]{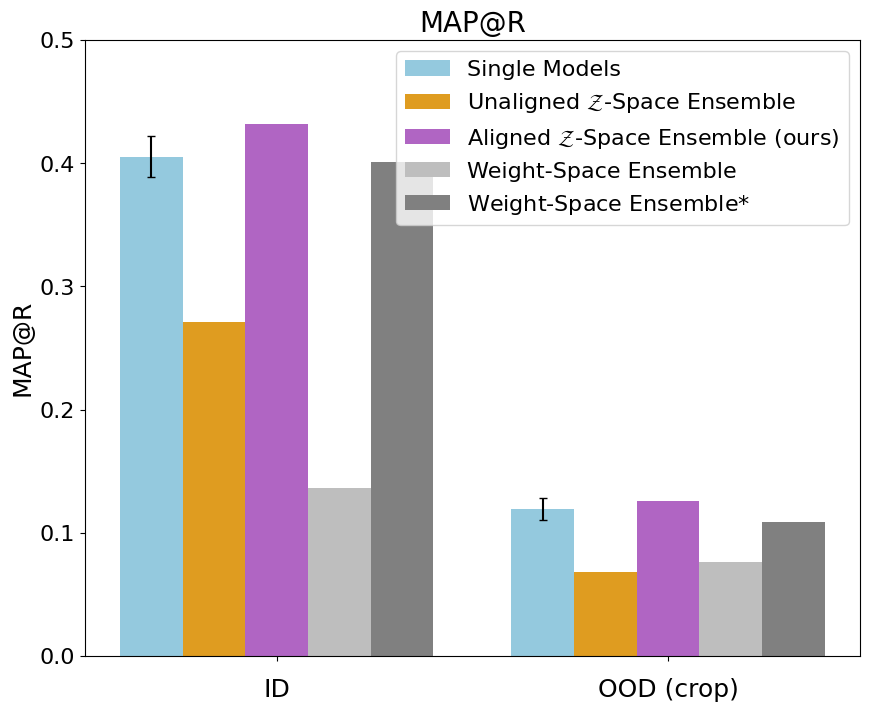} 
    \end{subfigure}
    \caption{Comparing embedding qualities of single models (blue), an ensemble of unaligned embedding spaces (orange), an ensemble of aligned embedding spaces (purple), a straightforward weight-space ensemble (WSE) (light grey), and a more constructed crafted weight-space ensemble (WSE*) \citep{rame2022diverse} (dark grey) in the ID and OOD settings. Recall@1 and MAP@R are presented. Higher values indicate better performance. The mean and standard deviation (error bars) of the performance metrics are reported for the 5 single models. The ensembles do not have standard deviation since all 5 models are combined into one. These plots show the same results as Figure \ref{fig:results-r1-mapr-crop-unsupervised}, but with the addition of the weight-space ensemble baselines.}
    \label{fig:results-r1-mapr-crop-unsupervised-diwa}
\end{figure}

\paragraph{Results} Figures \ref{fig:results-r1-mapr-color-unsupervised-diwa} and \ref{fig:results-r1-mapr-crop-unsupervised-diwa} compare the embedding quality across single models, unaligned embedding-space ensembles, aligned embedding-space ensembles, and both versions of weight-space ensemble discussed earlier. The straightforward weight-space ensemble approach (WSE), which uses the same single models as our embedding-space ensembles, consistently underperforms compared to the individual models, with performance reductions ranging from $36.13\%$ to $66.42\%$. The more carefully constructed weight-space ensemble (WSE*), which follows the standard weight-averaging practice, achieves performance comparable to the single models and consistently underperforms our aligned embedding-space ensemble in both ID and OOD settings. Detailed numerical values are provided in Tables \ref{tab:r1-mapr-single-models-vs-embedding-space-ensembles} and \ref{tab:r1-mapr-single-models-vs-weight-space-ensembles}.

\begin{table}[h!]
\small
    \centering
    \caption{Comparison of the embedding qualities of $M=5$ \textbf{single models}, unaligned embedding-space ensemble (\textbf{Unaligned Encoders}), and aligned embedding-space ensemble (\textbf{Aligned Encoders}) in both in-distribution (ID) and out-of-distribution (OOD) settings. The mean and standard deviation of performance metrics (Recall@1, MAP@R) are reported for the single models. Standard deviation is not shown for the ensembles since all 5 models are combined into one. The top three rows represent Recall@1 performance, and the bottom three rows represent MAP@R performance. The $\% \Delta$ column shows the percentage change in performance for each ensemble type relative to single models. All models are trained on the MNIST dataset using the InfoNCE contrastive loss, where positive pairs are created by applying random rotations to the same input image. }
    \vspace{1.5ex}
    \begin{tabular}{lccccccccc}
    \toprule
    \multirow{2}{*}{} & \multicolumn{2}{r}{\textbf{Single Models}} & \multicolumn{2}{c}{\textbf{Unaligned Encoders}} & \multicolumn{2}{c}{\textbf{Aligned Encoders}} \\
     & & {Mean $\pm$ Std}  & {Ensemble} & {\% $\Delta$} & {Ensemble} & {\% $\Delta$} \\
    \midrule
    \multirow{3}{*}{Recall@1 ($\uparrow$) } & ID & 0.900 $\pm$ 0.006 & 0.806 & -10.48\% & \textbf{0.911} & +1.18\% \\
     & OOD (Color) & 0.769 $\pm$ 0.029 & 0.664 & -13.68\% & \textbf{0.823} & +6.99\% \\
     & OOD (Crop) & 0.616 $\pm$ 0.011 & 0.525 & -14.80\% & \textbf{0.649} & +5.32\%  \\
    \midrule
    \multirow{3}{*}{MAP@R ($\uparrow$)} & ID & 0.405 $\pm$ 0.017 & 0.271 & -33.12\% & \textbf{0.432} & +6.61\% \\
     & OOD (Color) & 0.174 $\pm$ 0.012 & 0.128 & -26.35\% & \textbf{0.204} & +17.38\% \\
     & OOD (Crop) & 0.119 $\pm$ 0.009 & 0.068 & -42.95\% & \textbf{0.126} & +5.71\% \\
    \bottomrule
    \end{tabular}
    \label{tab:r1-mapr-single-models-vs-embedding-space-ensembles}
\end{table}

\begin{table}[h!]
\small
    \centering
    \caption{Comparing the embedding qualities of $M=5$ \textbf{single models}, straight-forward weight-space ensemble (\textbf{WSE}), and carefully constructed weight-space ensemble (\textbf{WSE*}) in the in-distribution (ID) and out-of-distribution (OOD) settings. The mean and standard deviation of the performance metrics (Recall@1, MAP@R) are reported for the single models. The ensembles do not have standard deviation since all 5 models are combined into one. Entries in the top three rows represent the Recall@1 performance, and entries in the bottom three rows represent the MAP@R performance. The $\% \Delta$ column shows the percentage change in the respective ensemble type compared to single models. The models are trained on the MNIST dataset using the InfoNCE contrastive loss, where positive pairs are created by applying random rotations to the same input image. }
    \vspace{1.5ex}
    \begin{tabular}{lccccccccc}
    \toprule
    \multirow{2}{*}{} & \multicolumn{2}{r}{\textbf{Single Models}} & \multicolumn{2}{c}{\textbf{WSE}} &  \multicolumn{2}{c} {\textbf{WSE*}}\\
     & & {Mean $\pm$ Std}  & {Ensemble} & {\% $\Delta$} & {Ensemble} & {\% $\Delta$} \\
    \midrule
    \multirow{3}{*}{Recall@1 ($\uparrow$) } & ID & 0.900 $\pm$ 0.006 & 0.544 & -39.56\% & 0.898 & -0.22\% \\
     & OOD (Color) & 0.769 $\pm$ 0.029 & 0.433 & -43.69\% & 0.771 & +0.26\% \\
     & OOD (Crop) & 0.616 $\pm$ 0.011 & 0.382 & -37.99\% & 0.610 & -0.97\% \\
    \midrule
    \multirow{3}{*}{MAP@R ($\uparrow$)} & ID & 0.405 $\pm$ 0.017 & 0.136 & -66.42\% & 0.401 & -0.99\% \\
     & OOD (Color) & 0.174 $\pm$ 0.012 & 0.068 & -60.92\% & 0.184 & +5.75\% \\
     & OOD (Crop) & 0.119 $\pm$ 0.009 &  0.076 & -36.13\% & 0.109 & -8.40\% \\
    \bottomrule
    \end{tabular}
    \label{tab:r1-mapr-single-models-vs-weight-space-ensembles}
\end{table}


\end{document}